\documentclass[10pt,twocolumn,letterpaper]{article}

\usepackage{cvpr}
\usepackage{times}
\usepackage{epsfig}
\usepackage{graphicx}
\usepackage{amsmath}
\usepackage{amssymb}
\usepackage{bm}
\usepackage{booktabs}
\usepackage{colortbl}
\definecolor{mygray}{gray}{.9}
\usepackage{enumitem}
\usepackage{adjustbox}
\usepackage{multirow}
\usepackage[OT1]{fontenc} 

\cvprfinalcopy 

\def\cvprPaperID{0473} 

\begin{document}

\title{Scene-based Factored Attention for Image Captioning}

\author{
  Chen Shen$^1$, Rongrong Ji$^{12}$\thanks{Corresponding author.}, Fuhai Chen$^1$, Xiaoshuai Sun$^1$, Xiangming Li$^1$\\
  $^1$Media Analytics and Computing Lab, Department of Artificial Intelligence, \\
  School of Informatics, Xiamen University, 361005, China.\\ 
  $^2$Peng Cheng Laboratory, Shenzhen, China. \\
  {\tt\small schenxmu@stu.xmu.edu.cn, rrj@xmu.edu.cn,} \\
  {\tt\small  \{cfh3c.xmu, xiaoshuaisun.hit\}@gmail.com, lixiangming@stu.xmu.edu.cn} 
}

\maketitle
\thispagestyle{empty}

\begin{abstract}
Image captioning has attracted ever-increasing research attention in the multimedia community. 
To this end, most cutting-edge works rely on an encoder-decoder framework with attention mechanisms, which have achieved remarkable progress. 
However, such a framework does not consider scene concepts to attend visual information, which leads to sentence bias in caption generation and defects the performance correspondingly.
We argue that such scene concepts capture higher-level visual semantics and serve as an important cue in describing images. 
In this paper, we propose a novel scene-based factored attention module for image captioning. 
Specifically, the proposed module first embeds the scene concepts into factored weights explicitly and attends the visual information extracted from the input image. Then, an adaptive LSTM is used to generate captions for specific scene types.
Experimental results on Microsoft COCO benchmark show that the proposed scene-based attention module improves model performance a lot, which outperforms the state-of-the-art approaches under various evaluation metrics.
\end{abstract}

\section{Introduction}
Describing what is in an image, known as image captioning, is a very challenging task, which attracts increasing attention in the multimedia research. 
In order to translate images to sentences, an encoder-decoder architecture is typically adopted for image captioning \cite{vinyals2015show, xu2015show, vinyals2017show}, which has achieved promising performance. 
Recent works in image captions prefer the usage of attention mechanism, which forces image captioning to dynamically focus on different regional features as needed, rather than being locked by a static image representation.
Since object-centered visual concepts have been proven to be effective in visual recognition \cite{parikh2011relative}, some captioning methods \cite{wu2016value,yao2017boosting,gan2017semantic} also prefer to selectively attend a set of detected object-centered visual concepts. 
These concepts are then combined into the hidden states of recurrent neural network (RNN) for dynamic caption generation.

\begin{figure}[t]
  \includegraphics[width=3.35in]{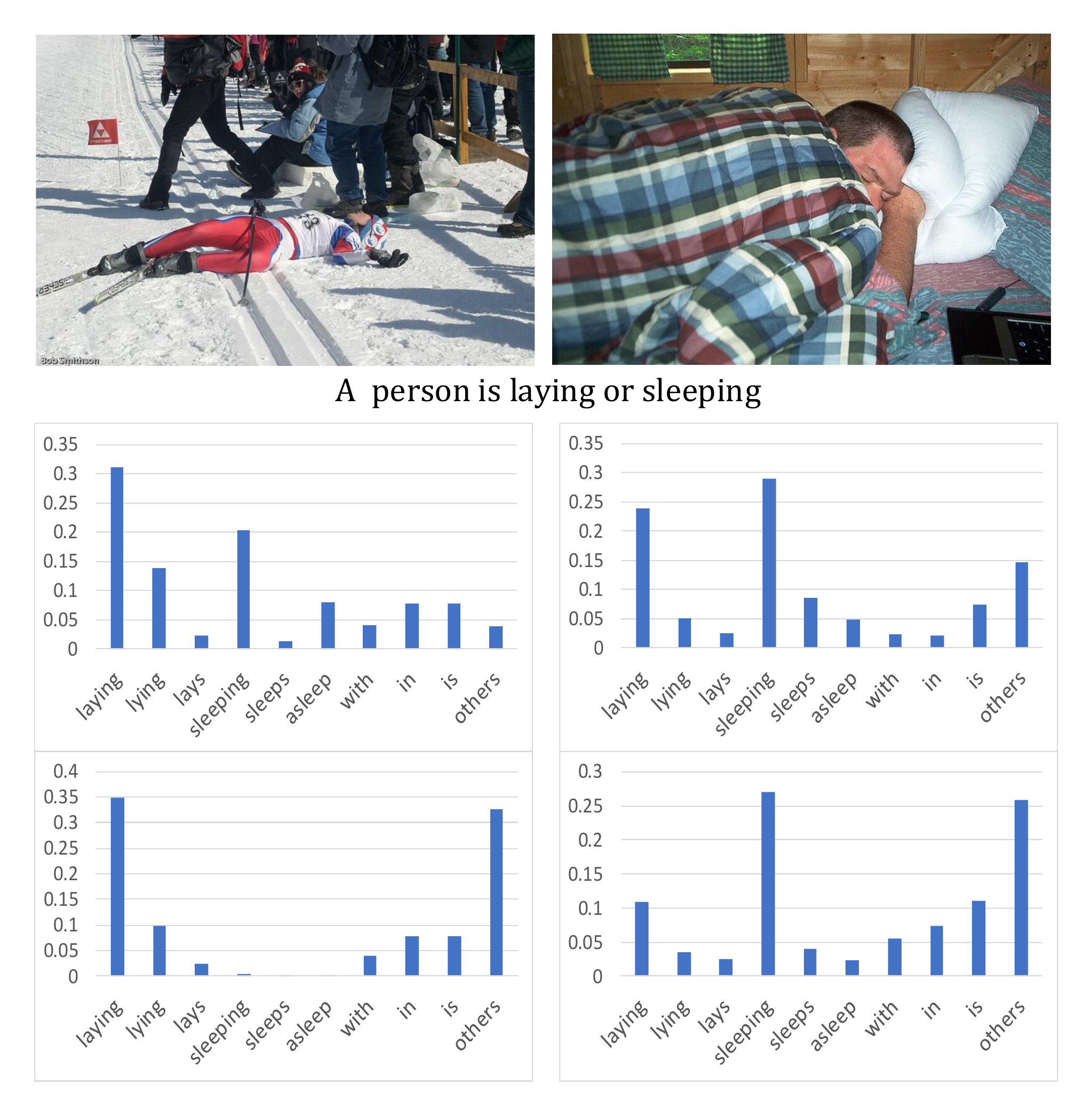}
  \caption{\textbf{Top}: Scene concepts affect word chosen in caption generation. \textbf{Middle}: Words probability distribution of leveraging scene concepts as semantic concepts. \textbf{Bottom}: Words probability distribution of our scene-based factored attention method.}
  \label{fig1}
\end{figure}

Despite the exciting recent progress, those works model attention based on either regional features or object-centered visual concepts. 
However, attention driven by scene concepts has never been explicitly considered, which actually plays a very important role in determining the major keywords of captions. 
As shown in the left case of Fig. \ref{fig1} (top), it is better to say "a person is laying\footnote{Due to the variance of crowdsourcing labeling, the word "laying" are used more frequently than "lying" in the captions of MS COCO dataset.}" than "a person is sleeping" when the scene is obviously outdoor. 
By contrast, in the right case of Fig. \ref{fig1} (top), when the photo is taken in a room with a man lying, it is more likely to get a caption as "a man is sleeping".
Clearly the scene concepts have a considerable influence on the word generation.

It is intuitive to introduce scene cues into image captioning.
A possible way to leverage the scene cues is to apply semantic concept attention. 
For example, one can follow You \emph{et al.} \cite{you2016image} to attend scene cues as semantic concepts for attention. 
Nevertheless, the visual information is always hierarchical \cite{li2012learning}, which makes the existing works suboptimal.
As the word probability distribution shown in Fig. \ref{fig1} (middle), after partial sentence generated for images in Fig. \ref{fig1} (top), the model with scene semantic attention is still not clear enough about choosing whether the word "laying" or "sleeping".
We argue that scene concepts and object-centered visual concepts should not be treated equally, since the scene concepts contain more global and macroscopic context information than object-centered visual concepts.
It therefore needs a more explicitly mechanism in the attention module as core guidance. 

\begin{figure*}[t]
  \centering
  \includegraphics[width=6in]{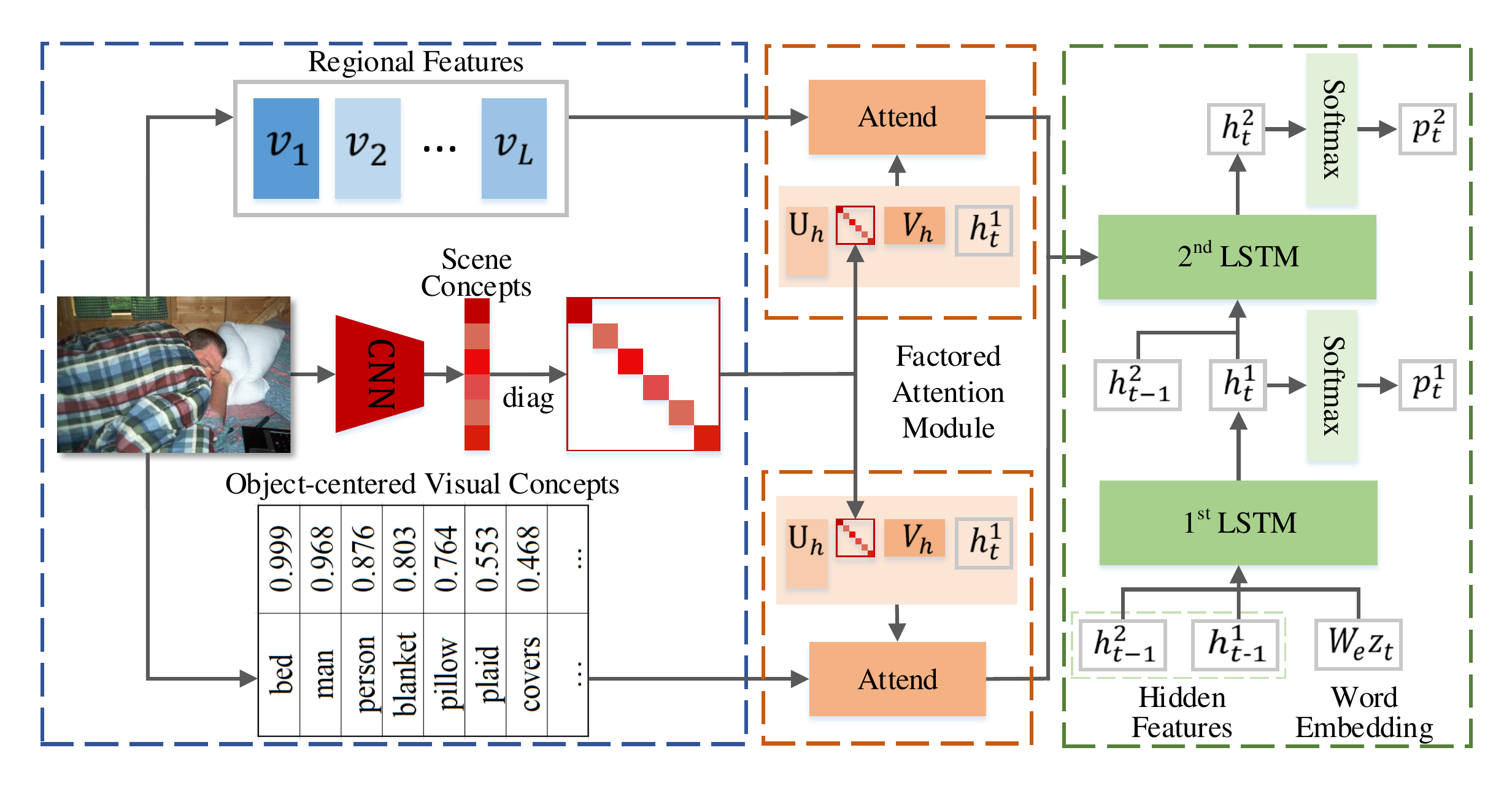}
  \caption{The overview of the proposed model. Given a set of visual information, \emph{i.e.}, regional features, object-centered visual concepts and scene concepts, which extracted from the input image in the encoder, factored attention module embeds scene concepts into the current hidden feature of the first LSTM to attend regional features and object-centered visual concepts. Then the weighted visual information is fed into the second LSTM to generate the next word in the Decoder.}
  \label{proposedModel}
\end{figure*}

In this paper, we argue that the fundamental issue lies in explicit and respective  modeling of scene concepts, object-centered visual concepts and sentence generation.
On one hand, the scene concepts are usually corresponding to the attribute keywords in captions. 
On the other hand, the context of scene concepts can guide to attend object-centered visual concepts when a sentence is generated. 
Driven by the above insights, we propose a novel scene-based factored attention module for image captioning.
The framework of the proposed method is illustrated in Fig. \ref{proposedModel}. 
To fully encode the input image, we first integrate the hierarchical visual information (including regional features, object-centered visual concepts and scene concepts) to enrich keywords and details in caption generation.
Then, we design a scene-based factored attention module to attend the hierarchical visual information.
Generally speaking, we embed scene concepts into the hidden feature of an LSTM \cite{hochreiter1997long}. Conditioned on the embedded scene hidden feature, the module determines which features and object-centered visual concepts are more important by assigning the corresponding weights. Finally, the outputs of the factored attention module are fed into a second LSTM to generate the next word.
As shown in Fig. \ref{fig1} (bottom), our model with scene-based factored attention is more confident with the chosen words.

The contributions of this paper are summarized as follows: (1) We are the first to explicitly embed scene concepts in image captioning. We are also the first to explicitly model relevance among scene concepts, object-centered visual concepts and caption generation. (2) We propose a factored attention module to better perceive the hierarchical visual information. Quantitative comparisons to the state-of-the-art demonstrate our merits.

\section{Related Work}
Our work relates to three topics: image captioning, tensor factorization and  scene understanding. In this section, we categorize and review related work as follows.

\subsection{Image Captioning}
Most existing image captioning methods rely on the encoder-decoder framework inspired by machine translation \cite{bahdanau2014neural,sutskever2014sequence}. The framework is used to "translate" an image to a sentence, where the visual features are extracted from convolutional neural network (CNN) and fed into Long Short-Term Memory (LSTM) to generate captions. Image captioning techniques have been extensively explored in \cite{karpathy2015deep,vinyals2015show,mao2014deep,chen2015mind,chen2017structcap,chen2018groupcap}. 
A few models \cite{xu2015show, you2016image, anderson2018bottom} seek to apply attention mechanism to bridge the gap of visual understanding and language processing. 
The prior attention mechanism relies on either regional convolution features or object-centered visual concepts extracted from images. 
The former allows the model to dynamically select regional features during sentence generation. 
And the latter, such as semantic attention \cite{you2016image, wu2016value}, applies top-down attention on detected object-centered visual concepts. 
However, these object-centered visual concepts have two major drawbacks. 
Firstly, they do not retain spatial information and scene guidance, which may make captions miss scene keywords and scene details.
Secondly, they do not take the hierarchy of semantics into account, which may lead to sentence bias.
As demonstrated in our experiments, considering the hierarchical semantic concepts at scene and object levels can better guide the attention selection and caption generation.

\subsection{Tensor Factorization}
Tensor factorization has been used in many multimedia tasks, such as attributes learning \cite{memisevic2007unsupervised}, motion style modeling \cite{taylor2009factored}, image
transformations \cite{sutskever2011generating} and sequence learning \cite{song2016factored, wu2016multiplicative}. 
Recently, tensor factorization has been widely used in \cite{kiros2014multimodal, fu2017aligning, gan2017semantic, gan2017stylenet}, which can further improve the model performance. More specifically, Kiros \emph{et al.} \cite{kiros2014multimodal} used factored tensor to guide word embedding with visual features. Fu \emph{et al.} \cite{fu2017aligning} inferred a topic vector (named scene vector) for tensor factorization in LSTM. Gan \emph{et al.} \cite{gan2017semantic} used factorization to remedy dimension explosion. Gan \emph{et al.} \cite{gan2017stylenet} introduced factored LSTM to learn different style captions. In contrast to these works, we use tensor factorization not only to explicitly model the relevance among visual information and sentence generation, but also to guide the attention selection mechanism.

\subsection{Scene understanding}
In the last few years, CNNs have emerged as powerful image representations for scene classification \cite{oliva2001modeling,wu2010semantics,song2010biologically,li2010object,yu2013pairwise,wang2017knowledge,guo2017locally}. Thanks to the development of Scene-15, MIT Indoor-67, SUN-397 and Place datasets \cite{zhou2014learning}, the well-known scene classification task has been pushed forward with great progress and gradually weeded out hand-crafted features. Recently, deep convolutional networks have been exploited for scene classification by Zhou \emph{et al.} \cite{zhou2014learning}.
We take full advantage of the recent scene understanding methods to help improve the quality of caption generation.

\section{The Proposed Model}
Firstly, a set of hierarchical visual information, \emph{i.e.}, regional features $v_{conv}$, object-centered visual concepts $v_{obj}$ and scene concepts $v_{scene}$ are extracted from the input image. 
Secondly, scene-based factored attention module embeds scene concepts $v_{scene}$ into the current hidden feature $h^1_t$ of the first LSTM to attend regional features $v_{conv}$ and object-centered visual concepts $v_{obj}$. 
Finally, the weighted visual information is fed into the second LSTM to generate the next word.

In Sec. \ref{Txt3:CaptionGeneration}, we briefly introduce the basic architecture of our proposed image captioning method. 
Then in Sec. \ref{Txt3:FactoredAttentionModule}, we introduce the factored attention module in details. 
Finally, in Sec. \ref{Txt3:objective}, we introduce the objective function used in our work.

\subsection{Caption Generation}
\label{Txt3:CaptionGeneration}
Long Short-Term Memory (LSTM) \cite{hochreiter1997long} is a widely-used Recurrent Neural Network (RNN), which is known to learn patterns with long-term temporal dependencies. 
We briefly refer to the operation of the LSTM over a single time step using the following notation:
\begin{equation}
  h_t = LSTM(x_t, h_{t-1}),\label{eqa:LSTM}
\end{equation}
where $x_t$ is the input vector of LSTM, and $h_t$ is the hidden feature of LSTM.

The hidden feature at time step $t$ can be calculated via Eq. \ref{eqa:LSTM}, formulated as follows:
\begin{align}
  i_t &= \sigma(W_ix_t + U_ih_{t-1}),\\
  f_t &= \sigma(W_fx_t + U_fh_{t-1}),\\
  o_t &= \sigma(W_ox_t + U_oh_{t-1}),\\
  g_t &= tanh(W_cx_t + U_ch_{t-1}),\\
  m_t &= f_t\odot m_{t-1} + i_t\odot g_t,\\
  h_t &= o_t\odot tanh(m_t),
\end{align}
where $i_t, f_t, o_t, m_t$ and $h_t$ are input gate, forget gate, output gate, memory cell and hidden feature, respectively. $\sigma$ and $\odot$ denote sigmoid function and an element-wise Hadamard product operator, respectively. For brevity, we omit all bias terms in the following paper.

LSTM's core is a memory cell $m_t$ that maintains the multi-modal knowledge of the inputs $x_t$ observed with respect to the time step $t$. 
Updating operations on the memory cell $m_t$ is modulated by three gates, \emph{i.e.}, the input gate $i_t$, the output gate $o_t$ and the forget gate $f_t$, which determine when and how the information flow. 
Especially, the input gate $i_t$ controls the input of the LSTM. 
The output gate $o_t$ manages the memory $m_t$ transfer to the hidden feature $h_t$ of the LSTM and generate the next word.
And the forget gate $f_t$ decides whether to forget previous memory $m_{t-1}$.

Our captioning model consists of two LSTM layers, referred as first LSTM and second LSTM. 
The superscript of variables in the equations is to distinguish which layer of LSTM.
The first LSTM generates a hidden feature of the current sequence $h^1_t$ based on the input, which contains partial sequence output generated so far, the current input word and the context information of the second LSTM. It is formulated as follows:
\begin{align}
  x^1_t &= [W_ez_t, h^2_{t-1}],\\
  h^1_t &= LSTM(x^1_t, h^1_{t-1}),
\end{align}
where $W_e\in \mathbb{R}^{E\times Q}$ is a word embedding matrix for a vocabulary of size $Q$. $z_t$ is the input word of a one-hot vector at time step $t$. 

We define the notation $y_{1:T}$ as a sequence of words $(y_1, y_2, ..., y_T)$, and get the first words conditional probability distribution at time step $t$ as follows:
\begin{equation}
  p^1(y_{t}|y_{1:t-1}) = Softmax(W_{y}h^1_t),
\end{equation}
where $W_{y} \in \mathbb{R}^{Q \times H}$ is a learned weight matrix.
Note that the output $p^1(y_{t}|y_{1:t-1})$ is a distribution of words only for loss optimization in training. The details will be described in Sec. \ref{Txt3:objective}.

In our proposed scene-based factored attention module, at each time step $t$, we use the current hidden feature $h^1_t$ to get the attentive weighted visual information $\hat{v}_t$, where the details will be described in Sec. \ref{Txt3:FactoredAttentionModule}.

We devise the second LSTM layer to make use of weighed visual information $\hat{v}_t$ to generate a word at each time step $t$, which can be further reformulated as:
\begin{align}
  x^2_t &= [\hat{v}_t, h^1_t],\label{eqa:2ndlstm1}\\
  h^2_t &= LSTM(x^2_t, h^2_{t-1}),\label{eqa:2ndlstm2}\\
  p^2(y_{t}|y_{1:t-1}) &= Softmax(W_{y}h^2_t),
\end{align}
where $W_{y} \in \mathbb{R}^{Q \times H}$ is a learned weight matrix.
The output $p^2(y_{t}|y_{1:t-1})$ is the second distribution of words, which not only participates in loss optimization in training, but is used independently to sample word in testing. The distribution of the whole generated caption $y_{1:T}$ is calculated as the product of conditional distributions:
\begin{equation}
  p^2(y_{1:T}) = \prod_{t=1}^{T}p^2(y_t|y_{1:t-1}),
\end{equation}

\subsection{Scene-based Factored Attention Module}
\label{Txt3:FactoredAttentionModule}
In order to take full advantages of scene concepts and model hierarchical semantic concepts, we further propose a factorization method to embed scene concepts into the attention mechanism.

We firstly obtain diagonal matrix $S \in \mathbb{R}^{s\times s}$ by direct diagonalization of scene concepts $v_{scene} \in \mathbb{R}^{s}$.
Then this diagonal scene matrix $S$ is embedded into the LSTM hidden feature $h^1_t$ by factorizing the parameters $W_{h}$ in the traditional attention mechanism \cite{xu2015show, anderson2018bottom} into three matrices $U_{h}$, $S$, $V_{h}$:
\begin{align}
  S &= diag(v_{scene}),\\
  W_{h} &= U_{h} S V_{h},
\end{align}
where $ U_{h} \in \mathbb{R}^{M\times s}$ and  $V_{h}\in \mathbb{R}^{s \times H}$ are the learned weight matrices that shared by all images and scene concepts.

The factored $W_{h}$ is used to transform the hidden feature $h^1_t$, which fuels the context of the scene concepts directly.
Therefore, the hidden feature $h^1_t$ obtains the context of the scene in this way. 
Given the regional features $v_{conv}=\{v_1,...,v_L\},v_i\in\mathbb{R}^{C}$, we generate first normalized attention weight $\alpha_t$ as follows:
\begin{align}
  a_{i,t} &=  w^T_a tanh(W_{va}v_i +U_{h}S V_{h}h^1_t),\\
  \alpha_t &= Softmax(a_{t}),\label{eqa:alphaWeight}\\
  \hat{v}_{conv,t} &= \sum^L_{i=1} \alpha_{i,t} v_{i},
\end{align}
where $W_{va}\in \mathbb{R}^{H\times V}$ and $W_{a} \in \mathbb{R}^{H}$ are the learned weight matrices.

Similarly, given the object-centered visual concepts $v_{obj}\in\mathbb{R}^{V}$, the second normalized attention weight $\beta_t$ is generated as follows:
\begin{align}
  b_{t} &=  w^T_b tanh((W_{vb}V_{obj} +U_{h}S V_{h}h^1_t),\\
  \beta_t &= Softmax(b_{t}),\\
  \hat{v}_{obj,t} &= \beta_t v_{obj},
\end{align}
where $W_{vb}\in \mathbb{R}^{H\times V}$ and $W_b \in \mathbb{R}^{H}$ are the learned weight matrices.

Finally, the weighted regional features $\hat{v}_{conv,t}$ and the weighted object-centered visual concepts $\hat{v}_{obj,t}$ are concatenated via Eq. \ref{eqa:concatV} and fed into the second LSTM in Eq. \ref{eqa:2ndlstm1} and Eq. \ref{eqa:2ndlstm2}.
\begin{equation}
  \hat{v}_t = [\hat{v}_{conv,t}, \hat{v}_{obj,t}],\label{eqa:concatV}
\end{equation}

\subsection{Objective Function}
\label{Txt3:objective}
Given a target ground-truth sequence $\bar{y}_{1:T}$ and a model with parameters $\theta$, we minimize the following maximum likelihood estimation (MLE) loss:
\begin{equation}
  L_{MLE}(\theta)= -\sum_{t=1}^{T}logp(\bar{y}_t|\bar{y}_{1:t-1}),
\end{equation}
In order to regularize the first LSTM more directly, we calculate the loss for both LSTMs as:
\begin{align}
  \begin{split}
    L_{MLE}(\theta) =& \gamma \cdot L^1_{MLE}(\theta) + (1 - \gamma)\cdot L^2_{MLE}(\theta)\label{eqa:totalLoss},\\
    =& -\gamma \cdot \sum_{t=1}^{T}logp^1(\bar{y}_t|\bar{y}_{1:t-1})\\
    &-(1-\gamma)\cdot \sum_{t=1}^{T}logp^2(\bar{y}_t|\bar{y}_{1:t-1}),
  \end{split}
\end{align}
where $\gamma$ is the hyper-parameter between 0 and 1.

Finally, we also introduce the reinforcement learning (RL) method into our framework for fair comparison with recent RL-based works like \cite{rennie2017self, anderson2018bottom, chen2018boosted,luo2018discriminability, gao2019deliberate}.\footnote{It should be noted that our scene-based factored attention module can be broadly used in other RL-based methods or GAN-based methods \cite{dai2017towards, chen2017show}.}
We minimize the negative expected reward after MLE training:
\begin{equation}
  L_R(\theta) = - \mathbb{E}_{y^s_{1:T}\sim p^2}[r(y^s_{1:T})],
\end{equation}
where $y^s_{1:T}$ is a sampled caption and $r$ is the CIDEr \cite{vedantam2015cider} reward function. Similar negative expected reward function has been proven to be effective in other works \cite{he2012maximum, rennie2017self, anderson2018bottom}.

Following the Self-critical Sequence Training (SCST) \cite{rennie2017self}, the gradient of $L_R(\theta)$ can be approximated:
\begin{equation}
  \bigtriangledown_\theta L_R(\theta)\approx-(r(y^s_{1:T})-r(\hat{y}_{1:T}))\bigtriangledown_\theta logp^2(y^s_{1:T}).
\end{equation}
where $y^s_{1:T}$ is a sampled caption and $\hat{y}_{1:T}$ defines the baseline score obtained by greedily decoding.

\section{Experiments}
In this section, we conduct extensive experiments to validate the effectiveness of scene-based factored attention module.
In Section \ref{Txt4:ExperimentalSetup}, we briefly introduce the dataset, images and captions pre-processing, evaluation metrics used in the experiments and implement details. 
Next, in Section \ref{Txt4:AblationStudy}, we discuss the ablation study of the proposed model.
Then in Section \ref{Txt4:ComparingSOTA}, we compare and analyze the results of the proposed model with other state-of-the-art models on image captioning both offline and online.
Finally, in Section \ref{Txt4:Qualitative}, we qualitatively analyze our merits in details.

\subsection{Experimental Settings}
\label{Txt4:ExperimentalSetup}

\subsubsection{Dataset}
In this paper, we utilize the MS COCO dataset \cite{chen2015microsoft}, which has been far and wide used in image captioning training and evaluation. MS COCO dataset contains 123,827 images. Each image in the dataset is given at least five captions by different Amazon’s Mechanical Turk (AMT) workers. Following the Karpathy split\footnote{https://github.com/karpathy/neuraltalk} in \cite{karpathy2015deep}, we use a set of 113,287 images for training, 5K images for validation and 5K for testing.

\begin{table*}[t]
  \centering
  \caption{Ablation study results on MS COCO Karpathy test split. The notation of "VC" denotes that we add traditional visual concepts attention and the notation of "Scene" denotes that we add factored attention module. The notation of "[VC, Scene]" denotes that we concatenate the visual concepts and scene concepts like "VC".}
  \label{AblationStudy}
  \setlength{\tabcolsep}{3.5mm}{
    \begin{tabular}{l|cccccccc}
      \toprule
      Model& Bleu1& Bleu2& Bleu3& Bleu4& METEOR& ROUGE& CIDEr& SPICE\\
      \midrule
      Baseline& 0.764& 0.602& 0.460& 0.349& 0.269& 0.559& 1.088& 0.201 \\
      Baseline + VC& 0.765& 0.605& 0.468& 0.359& 0.274& 0.564& 1.131& 0.205\\
      Baseline + Scene& \textbf{0.776}& 0.616& 0.473& 0.359& 0.271& 0.568& 1.124& 0.205\\
      Baseline + [VC, Scene]& \textbf{0.776}& \textbf{0.618}& 0.476& 0.361& 0.272& 0.567& 1.132& 0.208\\
      Baseline + VC + Scene& \textbf{0.776}& \textbf{0.618}& \textbf{0.477}& \textbf{0.367}&\textbf{ 0.277}& \textbf{0.570}& \textbf{1.147}& \textbf{0.209}\\
      \bottomrule
  \end{tabular}}
\end{table*}
\begin{table*}[t]
  \centering
  \caption{Single-model image captioning performance on MS COCO Karpathy test split. Results are reported for models trained with standard MLE loss in Table (top) and RL-based methods in Table (bottom). The numbers in boldface are the best known results and underlined numbers are the result of the second.}
  \label{ComparingSOTA}
  \setlength{\tabcolsep}{3.7mm}{
    \begin{tabular}{l|cccccccc}
      \toprule
      Model& Bleu1& Bleu2& Bleu3& Bleu4& METEOR& ROUGE& CIDEr& SPICE\\
      \midrule
      NIC\cite{vinyals2015show}& 0.663& 0.423& 0.277& 0.183& 0.237& -& 0.855& -\\
      Soft-Attention\cite{xu2015show}& 0.707& 0.492& 0.344& 0.243& 0.239& -& -& -\\
      Hard-Attention\cite{xu2015show}& 0.718& 0.504& 0.357& 0.250& 0.230& -& -& -\\
      ATT\cite{you2016image} & 0.709& 0.537& 0.402& 0.304& 0.243& - & -& -\\
      LSTM-A5\cite{yao2017boosting}& 0.730& 0.565& 0.429& 0.325& 0.251& 0.538& 0.986& -\\
      ARNet\cite{chen2018regularizing}& 0.740& 0.576& 0.440& 0.335& 0.261& 0.546& 1.034& 0.190\\
      LTG-Review-Net\cite{jiang2018learning}& 0.743& 0.579& 0.442& 0.336& 0.261& 0.548& 1.039& -\\
      Up-Down\cite{anderson2018bottom}& 0.772& -& -& 0.362& 0.270& 0.564& 1.135& 0.203\\
      DA\cite{gao2019deliberate} & \underline{0.758}& -& -& \underline{0.357}& \underline{0.274}& \underline{0.562}& \underline{1.119}& \underline{0.205}\\
      Ours& \textbf{0.776}& \textbf{0.618}& \textbf{0.477}& \textbf{0.367}& \textbf{0.277}& \textbf{0.570}&\textbf{ 1.147}& \textbf{0.209}\\
      \hline
      SCST:Att2in\cite{rennie2017self}& -& -& -& 0.313& 0.260& 0.543& 1.013& - \\
      SCST:Att2all\cite{rennie2017self}& -& -& -& 0.300& 0.259& 0.534& 0.994& -\\
      BAM\cite{chen2018boosted}& -& -& -& 0.350& 0.262& 0.559& 1.111& -\\
      ATTN+C+D(1)\cite{luo2018discriminability}& -& -& -& 0.363& 0.273& 0.571& 1.141& 0.211\\
      Up-Down\cite{anderson2018bottom}& 0.798& -& -& 0.363& 0.277& 0.569& 1.201& 0.214\\
      DA\cite{gao2019deliberate}& \underline{0.799}& -& -& \underline{0.375}& \textbf{0.285}& \textbf{0.582}& \underline{1.256}& \textbf{0.223}\\
      Ours& \textbf{0.803}& \textbf{0.646}& \textbf{0.601}& \textbf{0.381}& \textbf{0.285}& \textbf{0.582}& \textbf{1.268}& \underline{0.220}\\
      \bottomrule
  \end{tabular}}
\end{table*}

\subsubsection{Images and Captions Pre-processing}
In the encoder-decoder framework, image encoder is an essential part of image captioning, which is used to extract the visual information of images. To totally understand the input image $I$, we design three different kinds of visual information with hierarchical visual levels.
The low-level is the region feature $V_{conv}=\{v_1,...,v_k\},v_i\in\mathbb{R}^{C}$ extracted from the output of a Faster R-CNN \cite{ren2015faster} with ResNet-101 \cite{he2016deep} like other methods in \cite{anderson2018bottom, luo2018discriminability, gao2019deliberate}. And note that the number of regional features varies from image to image.
The middle-level is the object-centered visual concepts $V_{obj}\in\mathbb{R}^{V}$, which extracted from a visual concept extractor CNN trained on MS COCO dataset \cite{chen2015microsoft}. We refer nouns from captions as our visual semantic concepts. We regard it as a multi-label classification problem by minimizing a label smoothing \cite{szegedy2016rethinking} element-wise logistic loss function.
The high-level is the scene concepts $V_{scene}\in\mathbb{R}^{S}$, which is extracted from a scene classifier CNN pretrained on Place dataset \cite{zhou2014learning}. 

We follow standard practice and perform only minimal text-precessing. All the sentences in the training set are truncated to 16 characters, converting all sentences to lower case, tokenizing on white space, and filtering words that do not occur at least 5 times, resulting in a model vocabulary of 9,487 words.

\subsubsection{Evaluation Metric}
To evaluate the quantitative performance of the captions generated by our proposed model, we used five metrics which are commonly used in image captioning, including BLEU \cite{papineni2002bleu}, METEOR \cite{denkowski2014meteor}, ROUGE \cite{lin2004rouge}, CIDEr \cite{vedantam2015cider} and SPICE \cite{anderson2016spice}.
All the results are evaluated by Microsoft COCO caption evaluation tool\footnote{https://github.com/tylin/coco-caption}, where a larger score number in the results means better performance for all five metrics.

\subsubsection{Implementation Details}
We set the number of hidden units in each LSTM to 1,000, the number of hidden units in the attention layer to 512, and the size of the input word embedding to 1,000.
In training, the Adam optimizer \cite{kinga2015method} with a learning rate initialized to 5e-4 and decay by a factor 0.8 for every three epochs. 
The batch size is 100.
In testing, beam search is used to sample captions and the beam size is set to 2.

\begin{table*}[t]
  \centering
  \caption{Quantitative comparisons to the state-of-the-art works in image captioning on dataset c5 and c40 evaluated on the online MS-COCO server. Both SCST:Att2all and Up-Down are an ensemble of 4 models while ours is a single model. LSTM-A3 utilizes Resnet-152 based visual feature. The numbers in bold are the best and the underlined numbers are the second.}
  \label{comparedOnline}
  \scalebox{0.9}{
  \setlength{\tabcolsep}{1.8mm}{
    \begin{tabular}{l|c|c|c|c|c|c|c|c|c|c|c|c|c|c}
      \hline
      \multicolumn{1}{l|}{\multirow{2}{*}{Model}} & \multicolumn{2}{c|}{Bleu1} & \multicolumn{2}{c|}{Bleu2} & \multicolumn{2}{c|}{Bleu3} & \multicolumn{2}{c|}{Bleu4} & \multicolumn{2}{c|}{METEOR} & \multicolumn{2}{c|}{ROUGE} & \multicolumn{2}{c}{CIDEr}\\
      \cline{2-15}
      &  c5 & c40 & c5 & c40 & c5 & c40 & c5 & c40 & c5 & c40 & c5 & c40 & c5 & c40\\
      \hline
      Google NIC\cite{vinyals2015show}& 0.713& 0.895& 0.542& 0.802& 0.407& 0.694& 0.309& 0.587& 0.254& 0.346& 0.530& 0.682& 0.943& 0.946\\
      ATT\cite{you2016image}&           0.731& 0.901& 0.565& 0.816& 0.424& 0.710& 0.316& 0.600& 0.251& 0.336& 0.535& 0.683& 0.944& 0.959\\
      Review Net\cite{yang2016review}&  0.720& 0.900& 0.550& 0.812& 0.414& 0.705& 0.311& 0.597& 0.256& 0.347& 0.535& 0.686& 0.965& 0.969\\
      Adaptive\cite{lu2017knowing}&     0.748& 0.920& 0.584& 0.845& 0.444& 0.744& 0.336& 0.637& 0.264& 0.359& 0.555& 0.705& 1.042& 1.059\\
      PG-BCMR\cite{Liu2017Improved}&    0.754& 0.918& 0.591& 0.841& 0.445& 0.738& 0.332& 0.624& 0.257& 0.340& 0.550& 0.695& 1.013& 1.031\\
      SCST:Att2all\cite{rennie2017self}&0.781& 0.937& 0.619& 0.860& 0.470& 0.759& 0.352& 0.645& 0.270& 0.355& 0.563& 0.707& 1.147& 1.167\\
      LSTM-A3\cite{yao2017boosting}&    0.787& 0.937& 0.627& 0.867& 0.476& 0.765& 0.356& 0.652& 0.270& 0.354& 0.564& 0.705& 1.160& 1.180\\
      DA\cite{gao2019deliberate}&       0.794& 0.944& 0.635& 0.880& 0.487& 0.784& 0.368& 0.674& \textbf{0.282}& \underline{0.370}& \underline{0.577}& 0.722& \underline{1.205}& \underline{1.220}\\
      Up-Down\cite{anderson2018bottom}& \underline{0.802}& \textbf{0.952}& \underline{0.641}& \textbf{0.888}& \underline{0.491}& \underline{0.794}& \underline{0.369}& \underline{0.685}& 0.276& 0.367& 0.571& \underline{0.724}& 1.179& 1.205\\
      Ours& \textbf{0.803}& \underline{0.947}& \textbf{0.647}& \underline{0.887}& \textbf{0.500}&\textbf{0.797}& \textbf{0.379}& \textbf{0.690}& \textbf{0.282}& \textbf{0.372}& \textbf{0.581}& \textbf{0.730}& \textbf{1.235}& \textbf{1.256}\\
      \hline
    \end{tabular}}}
\end{table*}

\subsection{Ablation Study}
\label{Txt4:AblationStudy}
In order to figure out the contribution of each component, we conduct the following ablation studies on the MS COCO dataset with Karpathy test split. Specifically, we remove the visual concepts (VC) and the proposed factored attention module (Scene) respectively from our model.

We summarized the experimental results in Tab. \ref{AblationStudy}. 
The baseline is a re-implementation of Up-Down method proposed in \cite{anderson2018bottom}. 
The notation of "VC" denotes that we add traditional visual concepts attention and the notation of "Scene" denotes that we add factored attention module.
The notation of "[VC, Scene]" denotes that we concatenate the visual concepts and scene concepts as semantic attention.
And the notation of "Baseline + VC + Scene" is our full model, which denotes that the baseline model with our scene-based factored attention module.

With the results in Tab. \ref{AblationStudy}, we can see that our model performs better than the baseline model with relative improvements range from 1.6\% to 6.3\%. 
With the guidance of scene concepts, the model makes better use of visual information. 
In addition, compared with "Baseline + [VC, Scene]", we can see that though adding scene cues in visual concepts attention helps the model choose words, this is not the optimal solution. "Baseline + VC + Scene" obtains higher performance on all 5 metrics.
This verifies the importance of our scene-based factored attention module. 

\begin{figure}[t]
  \includegraphics[width=3.4in]{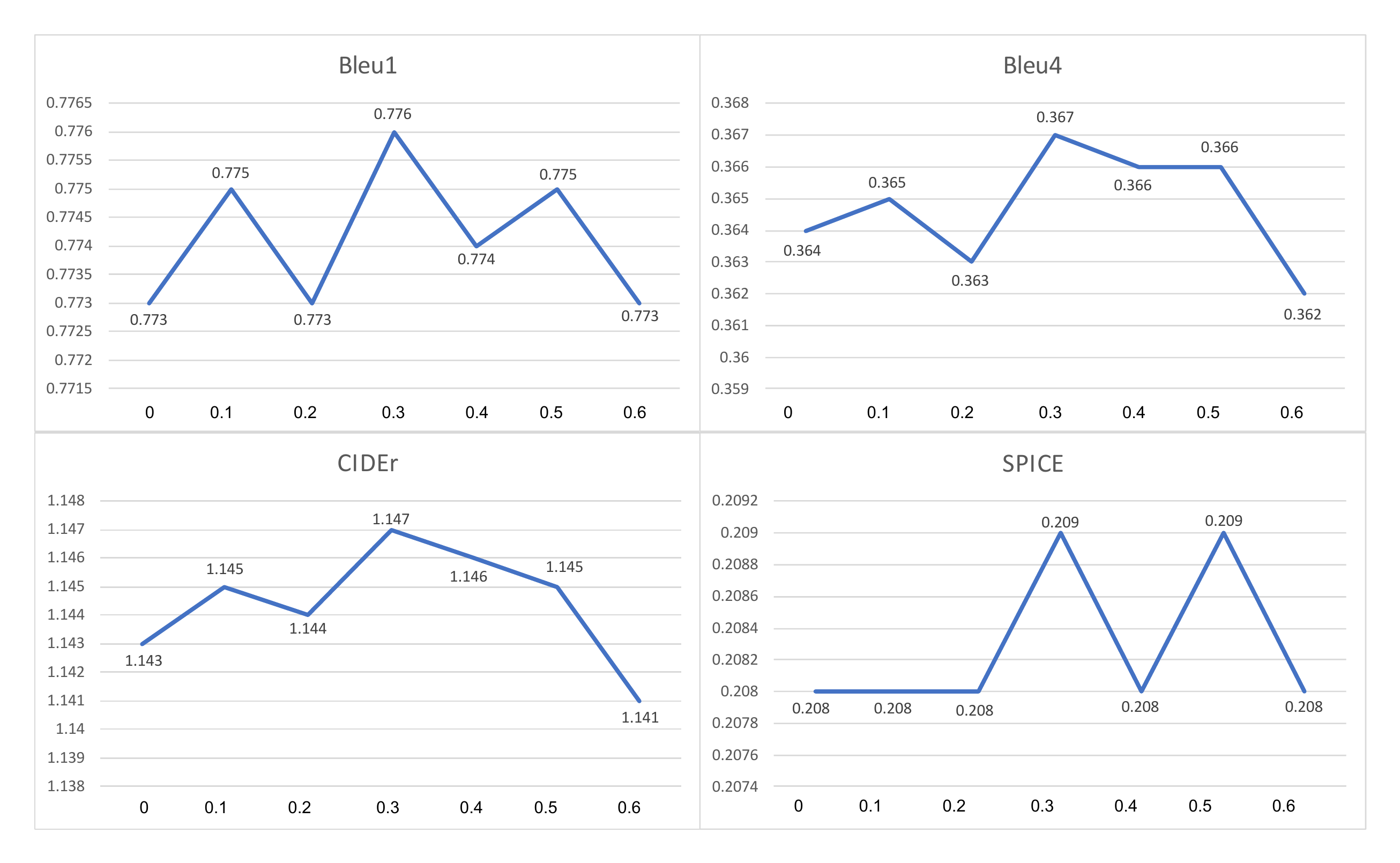}
  \caption{A variable-controlling experiment for $\gamma$ selection}
  \label{gammaSelect}
\end{figure}

In order to determine a hyper-parameters $\gamma$ as mentioned in the Eq. \ref{eqa:totalLoss}, we design an experiment with a variable-controlling approach. The objective results on the Karpathy test split with different $\gamma$ values are shown in Fig. \ref{gammaSelect}. Notice that evaluation results achieve their optimal scores when $\gamma=3$.

\subsection{Comparing with State-of-the-Arts}
\label{Txt4:ComparingSOTA}
In Tab. \ref{ComparingSOTA}, we report the performance of our framework in comparison to the existing state-of-the-arts on the test portion of the Karpathy splits.
For a fair comparison, results are reported for models trained with standard MLE loss in Tab. \ref{ComparingSOTA} (top), and models optimized for CIDEr score Tab. \ref{ComparingSOTA} (bottom).
For offline evaluation, all the image captioning models are single-model with no fine-tuning of the input ResNet / R-CNN model.
It is clear that our model performs the best on the generally used evaluation metrics, \emph{e.g.}, BLUE, ROUGE, CIDEr scores.
The experimental results demonstrate that our proposed scene-based factored attention module can significantly boost the scores compared with the existing state-of-the-arts

We also compare our model to the recent results on the official MS COCO evaluation by uploading results to the online MS COCO test server.
The online server provides "C5" and "C40" metrics which denote 5 reference captions and 40 reference captions, respectively.
The results are summarized in Tab. \ref{comparedOnline}, we can see that the performance of a single model trained with CIDEr optimization achieves the best performance on most metrics among the published state-of-the-art image captioning models on the blind test split.

\begin{figure*}[ht]
  \includegraphics[width=7in]{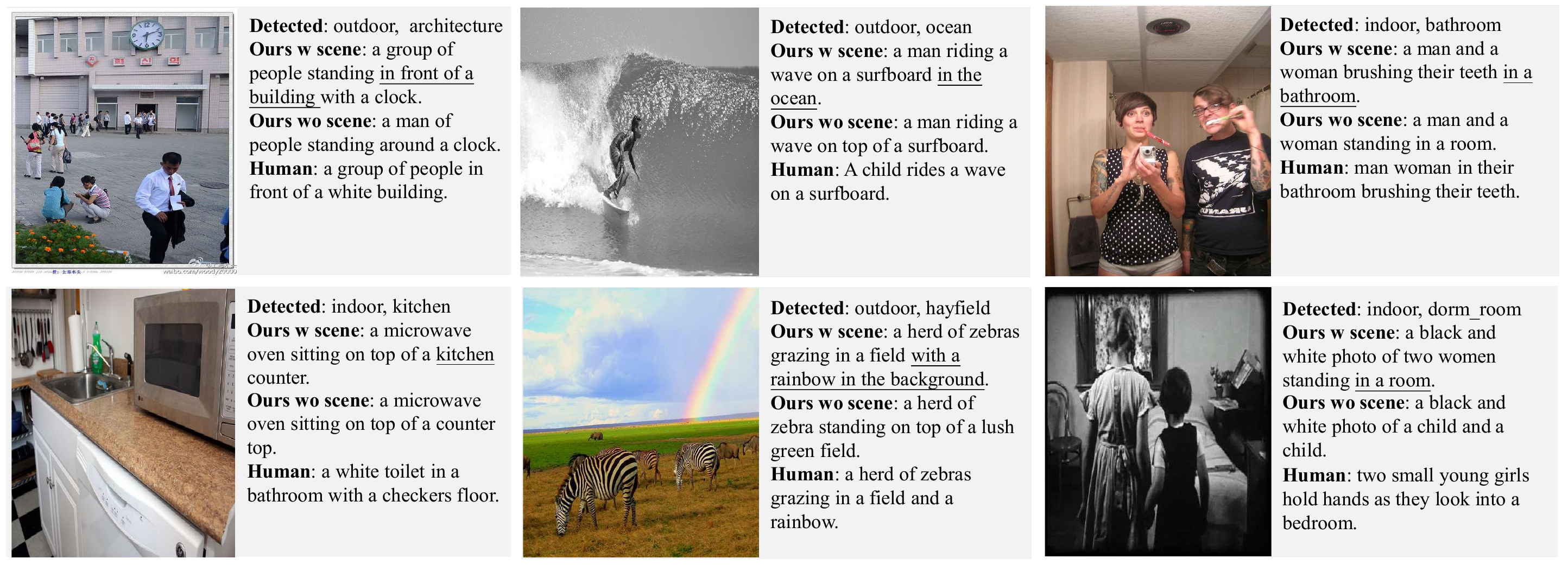}
  \caption{Qualitative analysis. The notation of "Detected" denote the scene concepts detected from the image. And the notations of "Ours w scene" and "Ours wo scene" denote our proposed model with/without scene-based factored attention module, respectively. It is easy to see that the model with the proposed module pays more attention to the details of the scenes, and the model is more inclined to mention the scene keywords in description generation.}
  \label{qualityAna}
\end{figure*}
\begin{figure*}[t]
  \centering
  \includegraphics[width=7in]{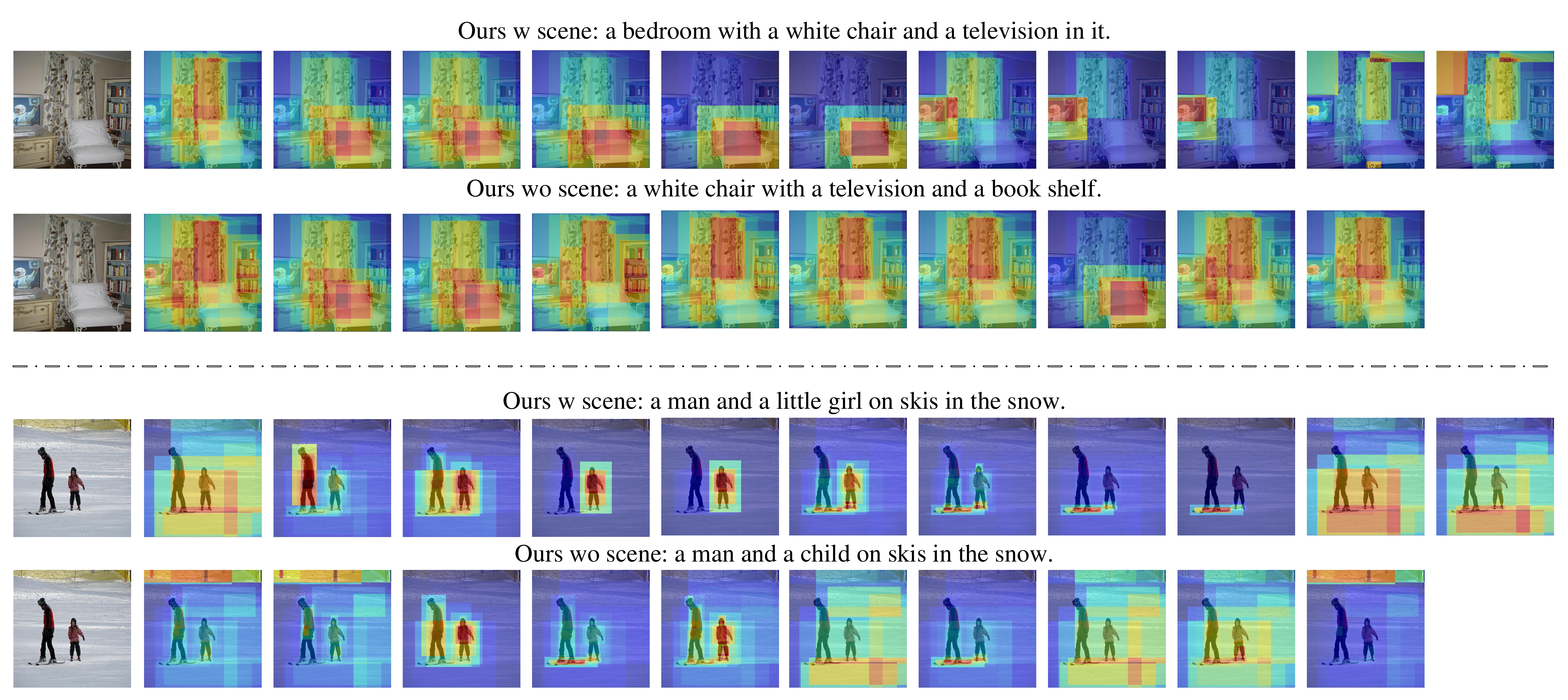}
  \caption{Visualization of attention regions with/without scene. The notations of "Ours w scene" and "Ours wo scene" denote our proposed model with/without scene-based factored attention module. The region with the maximum attention weight is in orange.}
  \label{attVisualization}
\end{figure*}

\subsection{Qualitative analysis}
\label{Txt4:Qualitative}
Here, we show some qualitative results in Fig. \ref{qualityAna} for a better understanding of our proposed model. 
The notation of "Detected" denote the scene concepts detected from the image. And notations of "Ours w scene" and "Ours wo scene" denote our proposed model with/without scene-based factored attention module. 
We can see that model with the proposed module pays more attention to the details of the scenes, and the proposed model is more inclined to mention the scene keywords in description generation.

We further visualize the heatmap of attention regions for words generated with/without scene-based factored attention module on the same image in Fig. \ref{attVisualization}. It is common practice \cite{xu2015show, anderson2018bottom} to directly visualize the attention weights $\alpha_t$ in Eq. \ref{eqa:alphaWeight} associated with word emitted at the same time step $t$.
We can find out that the area of attention is more clear with using the scene semantic concepts as guidance. In the complex scene as shown in the top of Fig. \ref{attVisualization}, it can pay more clearly and discriminately attention to regional features and tends to describe the scene more. 
In a relatively simple scene, as shown in the bottom of Fig. \ref{attVisualization}, the attention weights generated by our model are more logical, indicating that they are more accurate for the application of regional features of images.
As captions are being generated, the attention weights at both image examples vary properly when words generated.

\section{Conclusions}
In this work, we propose a novel scene-based factored attention module for image captioning. 
Different from previous works based on either regional features attention or object-centered visual concepts attention, our model takes scene concepts into account. As far as we know, we are the first to take scene concepts into consideration in image captioning and model relevance among scene concepts, object-centered visual concepts and caption generation. 
In our proposed scene-based factored attention module, we explicitly embed scene concepts in factored tensor into the LSTM hidden feature. Conditioned on the scene embedded hidden feature, we get the relative importance of regional features and object-centered visual concepts. 
The real power of our proposed module lies in its ability to attend hierarchically visual information for better captions.
Experiments conducted on the MS COCO captioning datasets validate the superiority of the proposed approach.

\section*{Acknowledgement}
This work is supported by the National Key R\&D Program (No.2017YFC0113000, and No.2016YFB1001503), Nature Science Foundation of China (No.U1705262, No.61772443, and No.61572410), Post Doctoral Innovative Talent Support Program under Grant BX201600094, China Post-Doctoral Science Foundation under Grant 2017M612134, Scientific Research Project of National Language Committee of China (Grant No. YB135-49), and Nature Science Foundation of Fujian Province, China (No. 2017J01125 and No. 2018J01106).

{\small
\bibliographystyle{ieee_fullname}
\bibliography{egpaper}
}

\end{document}